\documentclass{article}
\usepackage{spconf,amsmath,graphicx,amsfonts}
\usepackage{threeparttable}
\usepackage{booktabs} 
\usepackage{multirow}
\usepackage{color}
\usepackage{algorithm}
\usepackage{algorithmic}
\usepackage{verbatim}
\usepackage{arydshln}

\title{THERMAL INFRARED IMAGE INPAINTING VIA EDGE-AWARE GUIDANCE}
\name{$\text{Zeyu Wang}^a, \text{Haibin Shen}^a, \text{Changyou Men}^b, \text{Quan Sun}^b, \text{Kejie Huang}^{a,*}$
\thanks{This research has been supported by the China National Key R\&D Program (Grant No. 2022YFB4400704) and Hangzhou Major Technology Innovation Project of Artificial Intelligence (Grant No. 2022AIZD0060).}
\thanks{*Corresponding author: huangkejie@zju.edu.cn}}
\address{$^a$Zhejiang University, Hangzhou, China\\
$^b$Hangzhou Vango Technologies, Inc., Hangzhou, China}
\begin{document}
%
\maketitle
\begin{abstract}
Image inpainting has achieved fundamental advances with deep learning.
However, almost all existing inpainting methods aim to process natural images, while few target Thermal Infrared (TIR) images, which have widespread applications.
When applied to TIR images, conventional inpainting methods usually generate distorted or blurry content.
In this paper, we propose a novel task---\emph{Thermal Infrared Image Inpainting}, which aims to reconstruct missing regions of TIR images.
Crucially, we propose a novel deep-learning-based model \emph{TIR-Fill}.
We adopt the edge generator to complete the canny edges of broken TIR images.
The completed edges are projected to the normalization weights and biases to enhance edge awareness of the model.
In addition, a refinement network based on gated convolution is employed to improve TIR image consistency.
The experiments demonstrate that our method outperforms state-of-the-art image inpainting approaches on \emph{FLIR} thermal dataset.
\end{abstract}
\begin{keywords}
Thermal Infrared Image Inpainting, Edge Awareness, Gated Convolution, State-Of-The-Art.
\end{keywords}
\section{Introduction}
Image inpainting refers to filling missing regions of broken images, which has excellent usage in image processing.
Driven by the advances of Deep Learning (DL), this technique has progressed significantly in the past few years \cite{iizuka2017globally, liu2018image, yu2019free, nazeri2019edgeconnect, zheng2021tfill, wan2021high, zhu2021image}.
However, most inpainting methods aim to reconstruct natural images in the visible spectrum.

\begin{figure}[!tbp]
    \centering
    \includegraphics[width=1\linewidth]{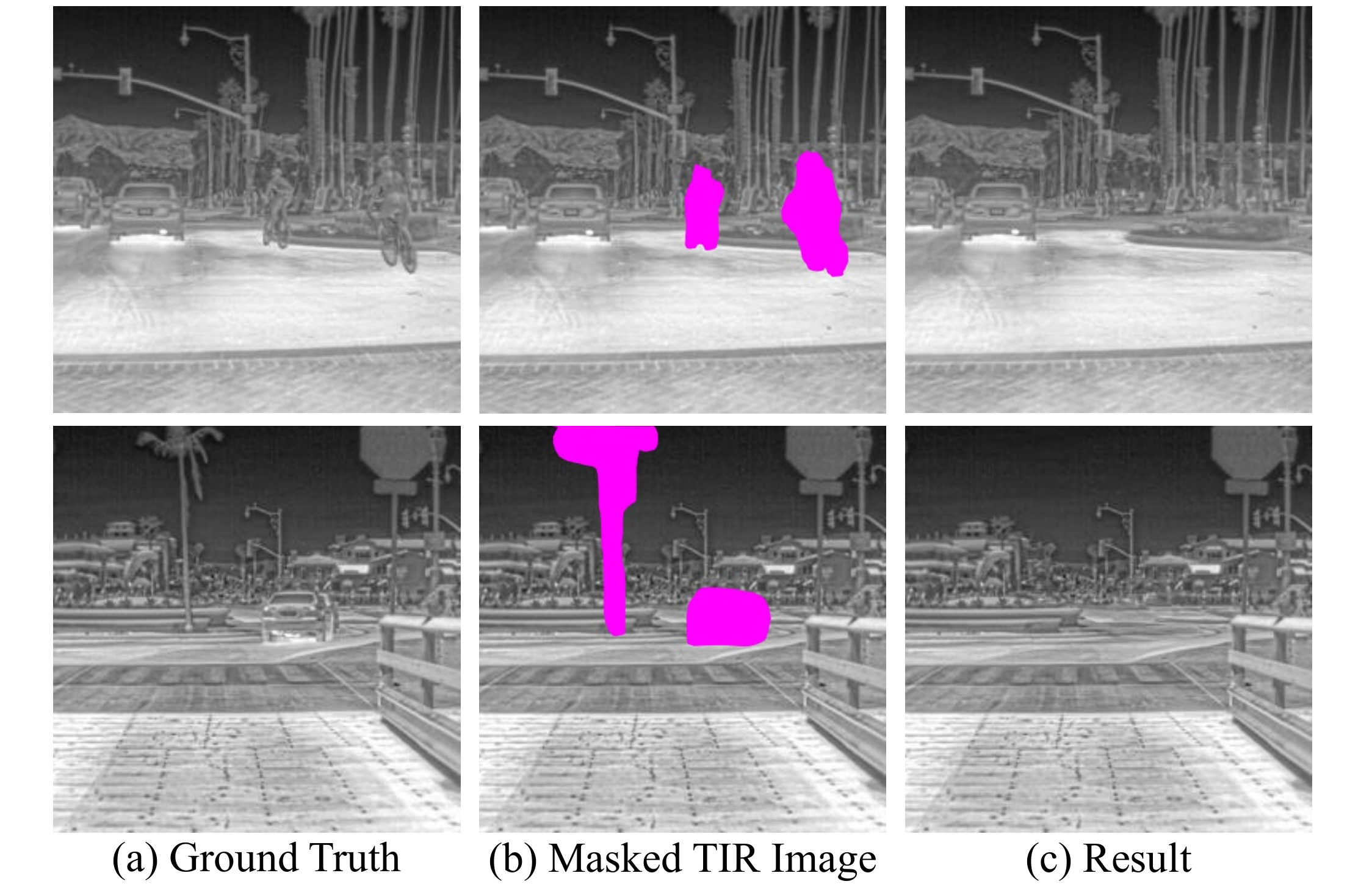}
    \caption{Examples of \emph{Thermal Infrared Image Inpainting}. The results are generated by our \emph{TIR-Fill}.}
    \label{Intro}
\end{figure}

Recently, Thermal Infrared (TIR) technology has been increasingly vital in remote sensing \cite{norman1995terminology}, medicine \cite{ring2012infrared}, and so on.
Unlike visible-spectrum cameras, TIR cameras can capture infrared radiation at the wavelength of $2\!-\!1000\mu m$, enabling them to detect low-light objects.
In this paper, we propose a novel task---\emph{Thermal Infrared Image Inpainting}, which aims to fill the missing regions of broken TIR images.
This technique can contribute to image editing, artifact repair, and object removal for TIR images, as shown in Fig. \ref{Intro}.

However, conventional DL-based inpainting methods are not applicable to TIR image inpainting, as they aim to reconstruct natural images but usually create distorted or blurry structures for TIR images.
Compared with natural images, TIR images have lower chromatic contrast but contain richer thermal information and sharper edge contours.
Therefore, we propose a novel DL-based model \emph{TIR-Fill}.
We first extract the canny edges of broken TIR images and build an edge generator to reconstruct them.
The completed edges are inserted into our Edge-Aware Guidance (EAG) normalization in \emph{TIR-Fill}, which enhances the model awareness of edge information.
A refinement network based on gated convolution is employed to improve the generated TIR image quality and consistency.
We demonstrate that our method quantitatively outperforms state-of-the-art image inpainting approaches and generates visually appealing results on \emph{FLIR} thermal dataset \cite{FLIR2019}, as shown in Fig. \ref{Intro}(c).


\begin{figure*}[!tbp]
    \centering
    \includegraphics[width=1\linewidth]{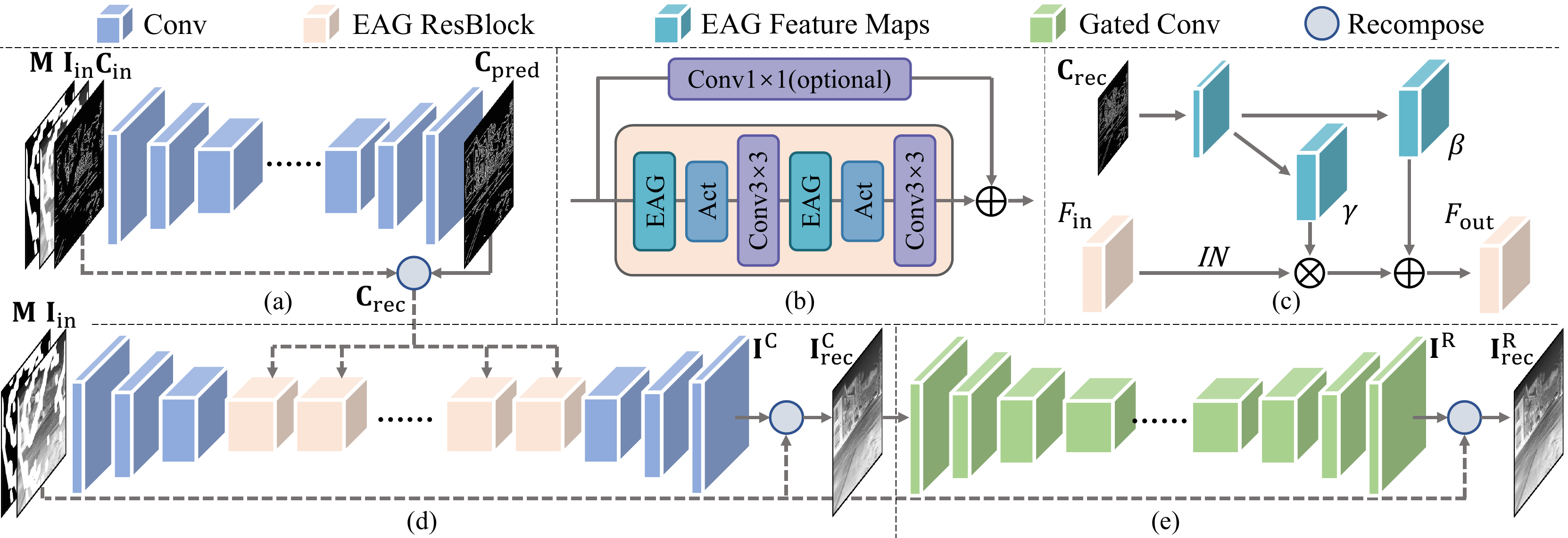}
    \caption{Illustration of our proposed \emph{TIR-Fill}. (a) The CNN-based edge generator $E_{\theta}$ \cite{nazeri2019edgeconnect}. (b) Our EAG ResBlock. (c) Our EAG normalization layer.
    (d) Our TIR image completion network $G_{\phi}$, which integrates EAG ResBlocks. (e) Our TIR image refinement network $R_{\psi }$ based on gated convolution \cite{yu2019free}.}
    \label{model}
\end{figure*}

\section{Related Works}

\subsection{Image Inpainting}
Traditional image inpainting methods include diffusion-based methods \cite{bertalmio2000image} and patch-based methods \cite{daisy2013fast}.
In recent years, Deep Learning (DL) has exhibited superior performance in image inpainting.
Conventional DL-based inpainting methods are based on Convolutional Neural Network (CNN) \cite{iizuka2017globally, liu2018image, yu2019free, nazeri2019edgeconnect, zhu2021image} and Transformer \cite{zheng2021tfill, wan2021high}.
The majority of these works are proposed for natural image inpainting.


\subsection{Thermal Infrared Image Processing}
TIR image processing has attracted extensive research for its vital and widespread applications.
Previous works targeted at TIR image super-resolution \cite{prajapati2021channel} and semantic segmentation \cite{ha2017mfnet}.
Currently, some works aim to tackle TIR image colorization, which maps a single-channel grayscale TIR image to an RGB image \cite{luo2021nighttime}.
To our knowledge, no DL-based work has been proposed for TIR image inpainting.

\begin{figure*}[!tbp]
    \centering
    \includegraphics[width=1\linewidth]{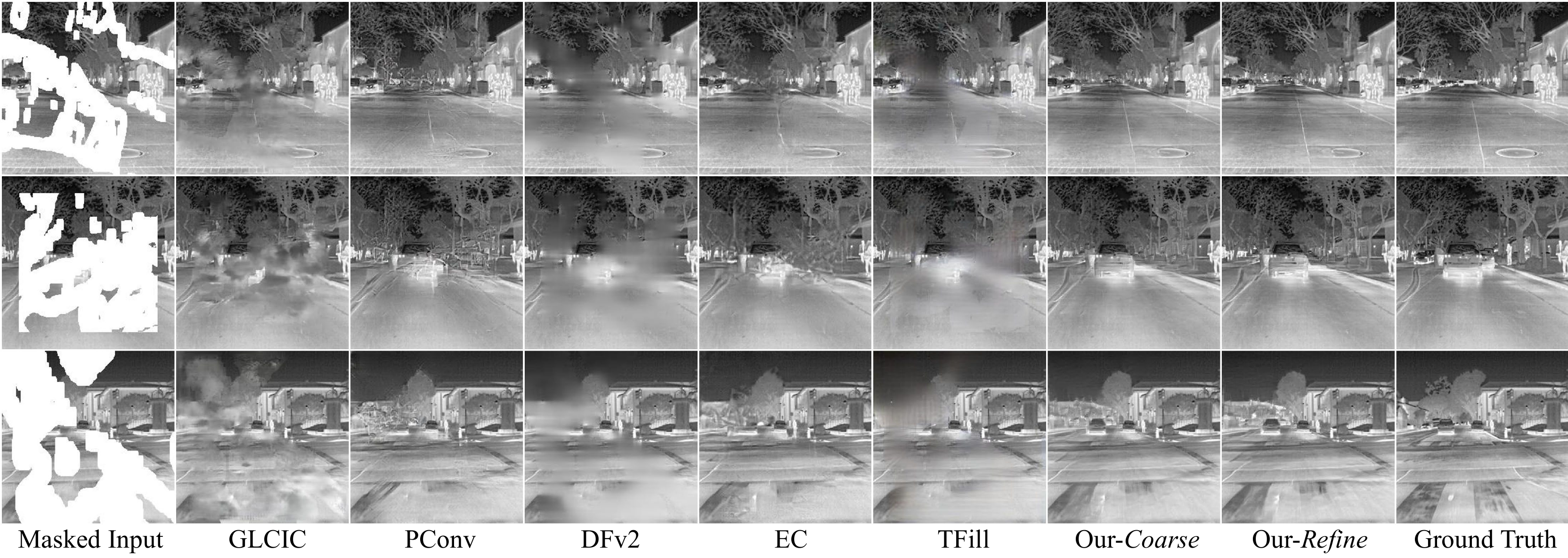}
    \caption{Qualitative comparison between our \emph{TIR-Fill} and the baseline inpainting methods on \emph{FLIR} dataset.}
    \label{qualitative}
\end{figure*}

\section{Methods}
\subsection{Overview}
To perform the task of \emph{Thermal Infrared Image Inpainting}, we propose the novel DL-based model \emph{TIR-Fill}, which is illustrated in Fig. \ref{model}.
It comprises three stages: Edge connection, TIR image completion, and TIR image refinement.
Given the mask $\mathbf{M}$, the goal of the task is to reconstruct the ground-truth TIR image $\mathbf{I}_{\text{gt}}$ based on the input $\mathbf{I}_{\text{in}}=\mathbf{I}_{\text{gt}}\odot\mathbf{M}$.
The mask $\mathbf{M}$ is a binary matrix, where 0 and 1 denote the hole and valid pixels, respectively.

\subsection{Edge Connection}
We first extract the canny edge $\mathbf{C}_{\text{in}}$ of the broken image $\mathbf{I}_{\text{in}}$.
The CNN-based edge generator $E_{\theta}$ \cite{nazeri2019edgeconnect} is adopted for the edge connection, as illustrated in Fig. \ref{model}(a).
The goal of this stage is to reconstruct the ground-truth edge $\mathbf{C}_{\text{gt}}$ based on $\mathbf{M}$, $\mathbf{I}_{\text{in}}$, and $\mathbf{C}_{\text{in}}$, as formulated below:
\begin{equation}
    \mathbf{C}_{\text{pred}}=\varepsilon (E_{\theta}(\mathbf{M}, \mathbf{I}_{\text{in}}, \mathbf{C}_{\text{in}}) - t_0 )
\end{equation}
where $t_0\!=\!0.5$ and $\varepsilon(\cdot )$ denotes the unit step function: if $t<t_0$, $\varepsilon(t-t_0)=0$, else $\varepsilon(t-t_0)=1$.

The predicted edge $\mathbf{C}_{\text{pred}}$ is recomposed with $\mathbf{C}_{\text{in}}$ into $\mathbf{C}_{\text{rec}} = \mathbf{C}_{\text{in}} + \mathbf{C}_{\text{pred}} \odot \mathbf{(1-M)}$, which will be adopted for enhancing the edge awareness.

\subsection{TIR Image Completion}
The TIR image completion network $G_{\phi}$ takes $\mathbf{M}$, $\mathbf{I}_{\text{in}}$, and $\mathbf{C}_{\text{rec}}$ as inputs to generate $\mathbf{I}^{\text{C}}$, as illustrated in Fig. \ref{model}(d).
\begin{equation}
    \mathbf{I}^{\text{C}}=G_{\phi}(\mathbf{M}, \mathbf{I}_{\text{in}}, \mathbf{C}_{\text{rec}})
\end{equation}

In detail, the intermediate layers of $G_{\phi}$ are our Edge-Aware Guidance (EAG) ResBlocks, as illustrated in Fig. \ref{model}(b).
Compared with conventional ResNet Block, EAG ResBlock replaces conventional normalization with our EAG normalization, which inserts the recomposed edge $\mathbf{C}_{\text{rec}}$, as shown in Fig. \ref{model}(c).
Inside the EAG normalization layer, $\mathbf{C}_{\text{rec}}$ is projected through convolutional layers into the modulation parameters $\gamma$ and $\beta$.
Then, $\gamma$ and $\beta$ are adopted as the normalization weight and bias, respectively, which is formulated as:
\begin{equation}
    F_{\text{out}} = \emph{IN}(F_{\text{in}}) \odot \gamma_{x, y, c}(\mathbf{C}_{\text{rec}}) + \beta_{x, y, c}(\mathbf{C}_{\text{rec}})
\end{equation}
where $\emph{IN}$ denotes the instance normalization, $\gamma = \gamma_{x, y, c}(\mathbf{C}_{\text{rec}})$, and $\beta = \beta_{x, y, c}(\mathbf{C}_{\text{rec}})$.
It is inspired from SPADE \cite{park2019semantic} normalization for semantic synthesis.
The predicted coarse result is recomposed into $\mathbf{I}_{\text{rec}}^{\text{C}} = \mathbf{I}_{\text{in}} + \mathbf{I}^{\text{C}} \odot \mathbf{(1-M)}$.

\subsection{TIR Image Refinement}
Based on coarse recomposed result $\mathbf{I}_{\text{rec}}^{\text{C}}$, a refinement network $R_{\psi }$ is employed to further improve the TIR image quality, as shown in Fig. \ref{model}(e).
\begin{equation}
    \mathbf{I}^{\text{R}}=R_{\psi }(\mathbf{M}, \mathbf{I}_{\text{rec}}^{\text{C}})
\end{equation}
The refinement network $R_{\psi }$ consists of stacked gated convolutional layers \cite{yu2019free}, which can learn dynamic feature gating for each channel and each spatial location:
\begin{equation}
    F_{\text{out}} = \sigma (W_{g} \ast F_{\text{in}}) \odot \phi (W_{f} \ast F_{\text{in}})
\end{equation}
where $W_{g}$ and $W_{f}$ denote two different convolutional filters, $\sigma$ denotes Sigmoid activation, and $\phi$ denotes Swish activation.
The predicted $\mathbf{I}^{\text{R}}$ is further recomposed with the input into the final result $\mathbf{I}_{\text{rec}}^{\text{R}} = \mathbf{I}_{\text{in}} + \mathbf{I}^{\text{R}} \odot \mathbf{(1-M)}$.

\subsection{Loss Functions}
To train $E_{\theta}$, we adopt adversarial training with Patch-GAN discriminator $D_{\text{patch}}$ \cite{zhu2017unpaired} and hinge loss.
\begin{equation}
    \mathcal{L}_{E_{\theta}}=\mathcal{L}_{\text{adv}}(\mathbf{C}_{\text{pred}})=-\mathbb{E}[D_{\text{patch}}(\mathbf{C}_{\text{pred}})]
    \label{adv}
\end{equation}

\begin{equation}
\begin{aligned}
\mathcal{L}_{D_{\text{patch}}}= &\mathbb{E}[relu(1-D_{\text{patch}}(\mathbf{C}_{\text{gt}}))] \\
+ &\mathbb{E}[relu(1+D_{\text{patch}}(\mathbf{C}_{\text{pred}}))]
\end{aligned}
\label{adv2}
\end{equation}

To train $G_{\phi}$ and $R_{\psi }$, we define a reconstruction loss consisting of $\ell_1$ loss, perceptual loss \cite{johnson2016perceptual}, and style loss \cite{gatys2016image}.
\begin{equation}
\begin{aligned}
    \mathcal{L}_{\text{rec}}(\mathbf{I}, \mathbf{I}_{\text{gt}}) = \ell_1(\mathbf{I}, \mathbf{I}_{\text{gt}}) + \sum\nolimits_{i}\|\mathcal{F}_{i}(\mathbf{I})-\mathcal{F}_{i}(\mathbf{I}_{\text{gt}})\|_{1} \\ + \sum\nolimits_{j}\|\mathcal{G}_{j}(\mathbf{I})-\mathcal{G}_{j}(\mathbf{I}_{\text{gt}})\|_{1}
\end{aligned}
\end{equation}
where $\mathcal{F}_{i}(i\!\in\! \{2, 7, 12, 21, 30\})$ denotes the intermediate feature map of the $i$-th layer in the VGG-19 \cite{Simonyan2015VeryDC}, 
and $\mathcal{G}_{j}(\cdot)\!=\!\mathcal{F}_{j}(\cdot) \mathcal{F}_{j}(\cdot)^{T} (j\!\in\! \{9, 18, 27, 32\})$ denotes the Gram matrix \cite{gatys2016image}.
Based on $\mathcal{L}_{\text{rec}}$, the loss functions of ${G_{\phi}}$ and ${R_{\psi}}$ are formulated as:
\begin{equation}
    \mathcal{L}_{G_{\phi}} = \mathcal{L}_{\text{rec}}(\mathbf{I}^{\text{C}}, \mathbf{I}_{\text{gt}})
\end{equation}
\begin{equation}
    \mathcal{L}_{R_{\psi}} = \mathcal{L}_{\text{rec}}(\mathbf{I}^{\text{R}}, \mathbf{I}_{\text{gt}}) + \mathcal{L}_{\text{adv}}(\mathbf{I}^{\text{R}})
\end{equation}
where another Patch-GAN discriminator is implemented, along with the adversarial loss $\mathcal{L}_{\text{adv}}(\mathbf{I}^{\text{R}})=-\mathbb{E}[D_{\text{patch}}(\mathbf{I}^{\text{R}})]$ and the same discriminator loss as Eq. \ref{adv2}.

\section{Experiments}

\subsection{Dataset}
We adopt \emph{FLIR} thermal dataset \cite{FLIR2019}, which consists of 8862 training images and 1366 testing images.
For training images, we randomly crop and resize them to $256 \times 256$.
For testing images, we resize them to $300 \times 375$ and crop them from the center to $256 \times 256$ for evaluation.
The irregular masks with arbitrary mask ratios provided by Liu \emph{et al.} \cite{liu2018image} are adopted.

\subsection{Experimental Settings}
We implement \emph{TIR-Fill} with PyTorch 1.8.1 on one NVIDIA RTX A6000 GPU with 40G memory.
The low and high thresholds of canny edge detection are set to 80 and 160, respectively.
The learning rates for training $E_{\theta}$, $G_{\phi}$, and $R_{\psi }$ are set to $1e^{-3}$, $1e^{-4}$, and $1e^{-4}$, respectively.
The Adam optimizer with $\beta_1=0.5$ and $\beta_2=0.9$ is employed.

\subsection{Quantitative Comparison}

We report PSNR, SSIM, LPIPS, and FID for quantitative comparison.
The results of our \emph{TIR-Fill} and the baseline inpainting methods are shown in Table \ref{quantitative}.
Among the baseline methods, DFv2 (DeepFillv2) performs well in PSNR, and even outperforms our coarse result under large mask ratios.
TFill is the SOTA method for natural image inpainting, but it performs poorly for TIR image inpainting.
By contrast, our \emph{TIR-Fill} achieves the best metrics under all mask ratios.
Obviously, our advantage in FID score is the most significant.

\begin{table}[!tbp]
    \small
    \setlength\tabcolsep{1.4pt} 
    \renewcommand\arraystretch{1.0} 
    \caption{Quantitative comparison between our \emph{TIR-Fill} and the baseline inpainting methods on \emph{FLIR} dataset.}
    \centering
    \begin{tabular}{c|c|c|c|c|c|c|c}
        \hline & Mask Ratio & 1-10 \!\% & 10-20 \!\% & 20-30 \!\% & 30-40 \!\% & 40-50 \!\% & 50-60 \!\% \\
        \hline \hline 
        \multirow{7}{*}{\rotatebox{90}{PSNR$\uparrow$}}
        & GLCIC \cite{iizuka2017globally}       & 36.61 & 31.02 & 27.53 & 25.11 & 23.04 & 20.28 \\
        & PConv \cite{liu2018image}             & 37.87 & 31.79 & 28.56 & 26.50 & 24.90 & 22.85 \\
        & DFv2 \cite{yu2019free}                & 37.82 & 32.24 & 29.20 & \textcolor{blue}{27.23} & \textcolor{blue}{25.72} & \textcolor{blue}{23.69} \\
        & EC  \cite{nazeri2019edgeconnect}      & 38.21 & 32.25 & 29.01 & 26.92 & 25.31 & 23.21 \\
        & TFill \cite{zheng2021tfill}           & 36.49 & 30.51 & 28.20 & 26.43 & 25.07 & 23.31 \\
        \cdashline{2-8}[3pt/2pt]
        & Our-\emph{Coarse}                     & \textcolor{blue}{39.36} & \textcolor{blue}{32.96} & \textcolor{blue}{29.45} & 27.13 & 25.38 & 23.17 \\
        & Our-\emph{Refine}                     & \textcolor{red}{39.65} & \textcolor{red}{33.48} & \textcolor{red}{30.06} & \textcolor{red}{27.79} & \textcolor{red}{26.06} & \textcolor{red}{23.82} \\
        \hline \hline 
        \multirow{7}{*}{\rotatebox{90}{SSIM$\uparrow$}}
        & GLCIC \cite{iizuka2017globally}       & 0.975 & 0.932 & 0.872 & 0.808 & 0.740 & 0.653 \\
        & PConv \cite{liu2018image}             & 0.977 & 0.935 & 0.881 & 0.824 & 0.766 & 0.682 \\
        & DFv2  \cite{yu2019free}               & 0.977 & 0.939 & 0.890 & 0.841 & 0.791 & \textcolor{blue}{0.727} \\
        & EC  \cite{nazeri2019edgeconnect}             & \textcolor{blue}{0.979} & 0.941 & 0.891 & 0.837 & 0.781 & 0.699 \\
        & TFill \cite{zheng2021tfill}           & 0.974 & 0.931 & 0.880 & 0.829 & 0.778 & 0.715 \\
        \cdashline{2-8}[3pt/2pt]
        & Our-\emph{Coarse}                     & \textcolor{red}{0.983} & \textcolor{blue}{0.950} & \textcolor{blue}{0.904} & \textcolor{blue}{0.853} & \textcolor{blue}{0.799} & 0.716 \\
        & Our-\emph{Refine}                     & \textcolor{red}{0.983} & \textcolor{red}{0.952} & \textcolor{red}{0.908} & \textcolor{red}{0.859} & \textcolor{red}{0.807} & \textcolor{red}{0.728} \\
        \hline \hline 
        \multirow{7}{*}{\rotatebox{90}{LPIPS$\downarrow$}}
        & GLCIC \cite{iizuka2017globally}       & 0.040 & 0.099 & 0.172 & 0.238 & 0.302 & 0.364 \\
        & PConv \cite{liu2018image}             & 0.029 & 0.074 & 0.127 & 0.177 & 0.228 & 0.290 \\
        & DFv2  \cite{yu2019free}               & 0.041 & 0.102 & 0.172 & 0.236 & 0.297 & 0.362 \\
        & EC  \cite{nazeri2019edgeconnect}             & \textcolor{blue}{0.028} & 0.072 & 0.123 & 0.173 & 0.224 & 0.289 \\
        & TFill \cite{zheng2021tfill}           & 0.038 & 0.094 & 0.154 & 0.208 & 0.263 & 0.322 \\
        \cdashline{2-8}[3pt/2pt]
        & Our-\emph{Coarse}                     & \textcolor{red}{0.021} & \textcolor{blue}{0.055} & \textcolor{blue}{0.097} & \textcolor{blue}{0.139} & \textcolor{blue}{0.184} & \textcolor{blue}{0.246} \\
        & Our-\emph{Refine}                     & \textcolor{red}{0.021} & \textcolor{red}{0.054} & \textcolor{red}{0.095} & \textcolor{red}{0.137} & \textcolor{red}{0.181} & \textcolor{red}{0.242} \\
        \hline \hline 
        \multirow{7}{*}{\rotatebox{90}{FID$\downarrow$}}
        & GLCIC \cite{iizuka2017globally}       & 8.41 & 26.27 & 54.78 & 83.82 & 114.85 & 145.63 \\
        & PConv \cite{liu2018image}             & 5.19 & 13.96 & 26.49 & 41.71 & 60.36 & 85.33 \\
        & DFv2  \cite{yu2019free}               & 8.23 & 26.15 & 54.24 & 85.10 & 115.96 & 141.67 \\
        & EC  \cite{nazeri2019edgeconnect}             & 4.78 & 12.39 & 23.10 & 35.83 & 52.02 & 80.48 \\
        & TFill \cite{zheng2021tfill}           & 9.41 & 25.97 & 50.13 & 77.49 & 108.41 & 136.61 \\
        \cdashline{2-8}[3pt/2pt]
        & Our-\emph{Coarse}                     & \textcolor{blue}{3.28} & \textcolor{blue}{8.12}  & \textcolor{blue}{14.29} & \textcolor{blue}{20.33} & \textcolor{blue}{27.36} & \textcolor{blue}{39.84} \\
        & Our-\emph{Refine}                     & \textcolor{red}{3.16} & \textcolor{red}{7.58}  & \textcolor{red}{12.86} & \textcolor{red}{17.76} & \textcolor{red}{23.13} & \textcolor{red}{31.86} \\
        \hline \hline 
        \end{tabular}
    \label{quantitative}
    \begin{tablenotes}
      \footnotesize
      \item $\uparrow$: Higher is better. $\downarrow$: Lower is better. 
      The values marked in \textcolor{red}{red} and \textcolor{blue}{blue} denote the best and the second best results, respectively.
    \end{tablenotes}
\end{table}

\subsection{Qualitative Comparison}
The qualitative results are shown in Fig. \ref{qualitative}.
We can see that these baseline methods generate poor inpainting results for TIR images.
DFv2 and TFill generate blurry content, while GLCIC and PConv create distorted structures.
EC (Edge Connector), which also utilizes edge information, can generate somewhat good results but with a few damaged details.
By contrast, our \emph{TIR-Fill} generates delicate structures, including pedestrians, trees, and cars.
The comparison illustrates that our model can generate visually appealing inpainting results for TIR images.
Therefore, the previous baseline methods are not applicable to TIR image inpainting, which is a worth-discussing task requiring the specifically designed method.

\subsection{Ablation Study}
\begin{figure}[!tbp]
    \centering
    \includegraphics[width=1\linewidth]{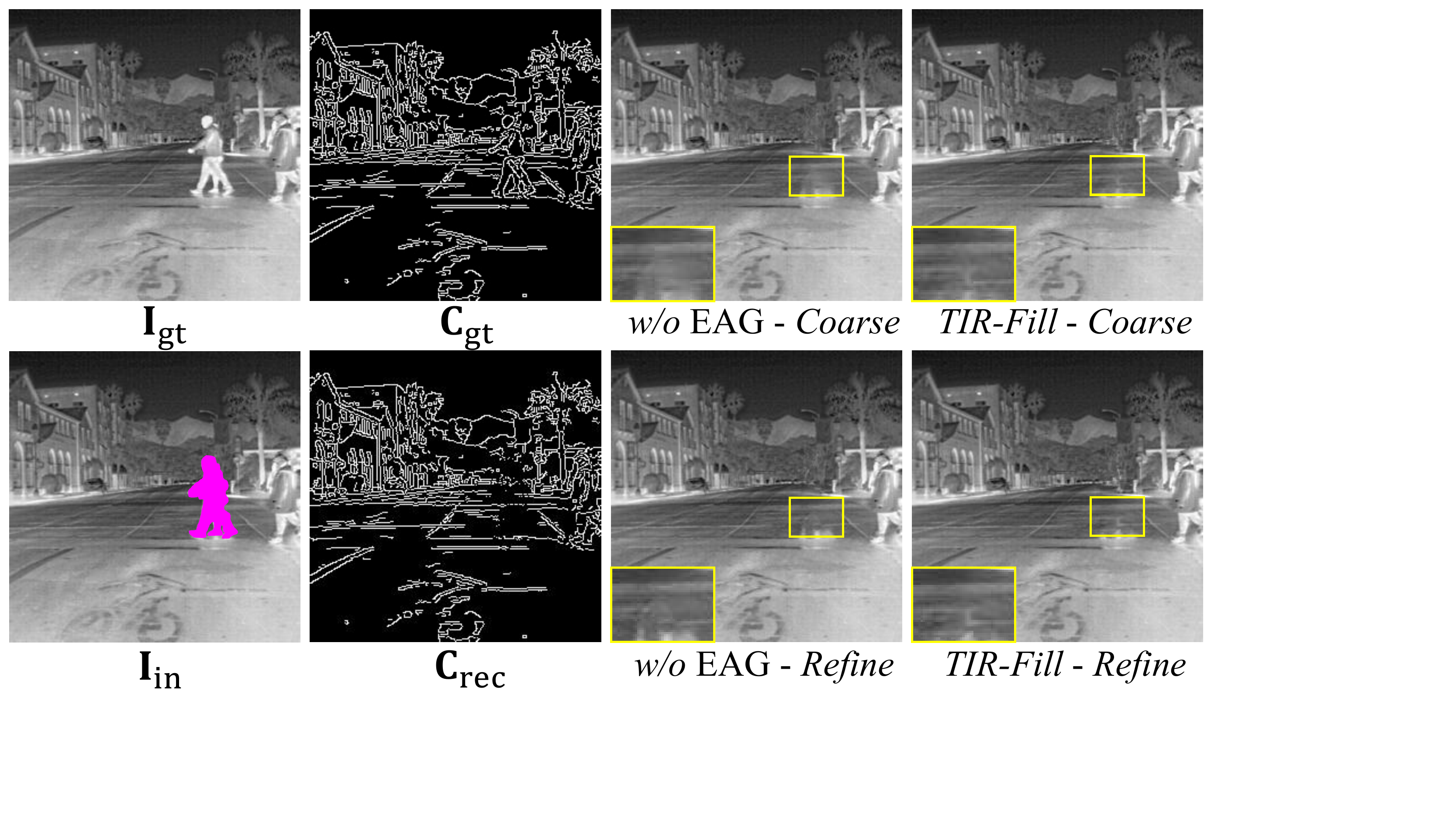}
    \caption{Qualitative results of the ablation study and visualization of the reconstructed canny edges.}
    \label{ablation_qualitative}
\end{figure}

We conduct the ablation study to illustrate the effectiveness of our EAG normalization.
The variant ``\emph{w/o} EAG'' denotes another \emph{TIR-Fill} which replaces the EAG normalization with conventional instance normalization.
The results shown in Table \ref{ablation_quantitative} demonstrate that EAG normalization dramatically improves the quantitative performance, especially the FID score.
In Fig. \ref{ablation_qualitative}, we aim to remove the pedestrians on the road.
The visualization of the reconstructed edge $\mathbf{C}_{\text{rec}}$ shows that the edges of the pedestrians are naturally removed and completed.
The qualitative comparison between the variant and the complete \emph{TIR-Fill} demonstrates that \emph{TIR-Fill} can generate more delicate structures.

\begin{table}[tbp]
    \small
    \setlength\tabcolsep{6pt} 
    \renewcommand\arraystretch{1.0} 
    \caption{Quantitative results of the ablation study. The results are shown as the average scores under all mask ratios.}
    \centering
    \begin{tabular}{c|c|c|c|c|c}
        \hline Model & Stage & PSNR$\uparrow$ & SSIM$\uparrow$ & LPIPS$\downarrow$ & FID$\downarrow$ \\
        \hline \hline 
        \emph{w/o} EAG  & \emph{Coarse}  & 29.24 & 0.844 & 0.135 & 25.61 \\
        \emph{w/o} EAG  & \emph{Refine}  & 29.73 & 0.849 & 0.133 & 21.42 \\
        \cdashline{1-6}[3pt/2pt]
        \emph{TIR-Fill} & \emph{Coarse}  & 29.57 & 0.867 & 0.124 & 18.87 \\
        \emph{TIR-Fill} & \emph{Refine}  & 30.14 & 0.873 & 0.122 & 16.06 \\
        \hline \hline 
        \end{tabular}
    \label{ablation_quantitative}
\end{table}

\section{Conclusion}
In this work, we propose a novel image processing task---\emph{Thermal Infrared Image Inpainting}, which aims to reconstruct missing regions of TIR images.
In addition, we propose an effective model \emph{TIR-Fill} to deal with the novel task, which integrates our EAG normalization for enhancing edge awareness.
The experiments demonstrate that our \emph{TIR-Fill} outperforms the baseline inpainting methods.
Its visually appealing inpainted results demonstrate its ability in TIR image editing.
The ablation study illustrates that EAG normalization performs better than conventional normalization.
In the future, we expect the worth-discussing task will attract extensive research and contribute to widespread applications.

\bibliographystyle{IEEEbib}
\bibliography{citation.bib}

\end{document}